\documentclass{bmvc2k}
\usepackage{multicol}

\title{{\fontsize{16}{16}\selectfont Delving Deeper into Anti-aliasing in ConvNets}}
\addauthor{Xueyan Zou}{zxyzou@ucdavis.edu}{1}
\addauthor{Fanyi Xiao}{fyxiao@ucdavis.edu}{1}
\addauthor{Zhiding Yu}{zhidingy@nvidia.com}{2}
\addauthor{Yong Jae Lee}{yongjaelee@ucdavis.edu}{1}

\addinstitution{
 University of California, Davis \\
 Davis, CA, USA
}
\addinstitution{
 NVIDIA \\
 Santa Clara, CA, USA
}

\runninghead{Zou, Xiao, Yu, Lee}{Delving Deeper into Anti-aliasing in ConvNets}

\newcommand{\colorsout}[1][\daniel]{\bgroup\markoverwith{#1{\rule[0.5ex]{2pt}{1.2pt}}}\ULon}

\usepackage[normalem]{ulem}
\usepackage{amsfonts}
\usepackage{multirow}
\usepackage{enumitem}
\usepackage[rightcaption]{sidecap}

\begin{document}
\maketitle


\begin{abstract}
Aliasing refers to the phenomenon that high frequency signals degenerate into completely different ones after sampling. It arises as a problem in the context of deep learning as downsampling layers are widely adopted in deep architectures to reduce parameters and computation. The standard solution is to apply a low-pass filter (e.g., Gaussian blur) before downsampling \cite{zhang2019making}. However, it can be suboptimal to apply the same filter across the entire content, as the frequency of feature maps can vary across both spatial locations and feature channels. To tackle this, we propose an adaptive content-aware low-pass filtering layer, which \emph{predicts separate filter weights for each spatial location and channel group} of the input feature maps. We investigate the effectiveness and generalization of the proposed method across multiple tasks including ImageNet classification, COCO instance segmentation, and Cityscapes semantic segmentation. Qualitative and quantitative results demonstrate that our approach effectively adapts to the different feature frequencies to avoid aliasing while preserving useful information for recognition. Code is available at \url{https://maureenzou.github.io/ddac/}.
\end{abstract}


\section{Introduction}

Deep neural networks have led to impressive breakthroughs in visual recognition, speech recognition, and natural language processing. On certain benchmarks such as ImageNet and SQuAD, they can even achieve ``human-level'' performance \cite{mnih-nature2015,he-iccv2015,tan2019efficientnet, rajpurkar2016squad}. However, common mistakes that these networks make are often quite \emph{unhuman} like. For example, a tiny shift in the input image can lead to drastic changes in the output prediction of convolutional neural networks (ConvNets)~\cite{shankar2019systematic, azulay2018deep, tan2019efficientnet}. This phenomenon was demonstrated to be partially due to \emph{aliasing} when downsampling in ConvNets~\cite{zhang2019making}.

Aliasing refers to the phenomenon that high frequency information in a signal is distorted during subsampling~\cite{gonzales2002digital}. The Nyquist theorem states that the sampling rate must be at least twice the highest frequency of the signal in order to prevent aliasing. Without proper anti-aliasing techniques, a subsampled signal can look completely different compared to its input. Below is a toy example demonstrating this problem on 1D signals:

\vspace{-20pt}
\begin{equation}
\begin{split}
001100110011 \xrightarrow[\text{maxpool}]{\text{k=2, stride=2}} 010101 \\
011001100110 \xrightarrow[\text{maxpool}]{\text{k=2, stride=2}} 111111
\end{split}
\label{eq:aliasing}
\end{equation}
\vspace{-4pt}

Here $k$ is the kernel size ($1\times2$). Because of aliasing, a one position shift in the original signal leads to a completely different sampled signal (bottom) compared to the original sampled one (top). As downsampling layers in ConvNets are critical for reducing parameters and inducing invariance in the learned representations, the aliasing issue accompanying these layers will likely result in a performance drop as well as undesired shift variance in the output if not handled carefully.


\begin{figure*}[t]
\begin{center}
\includegraphics[width=.89\textwidth]{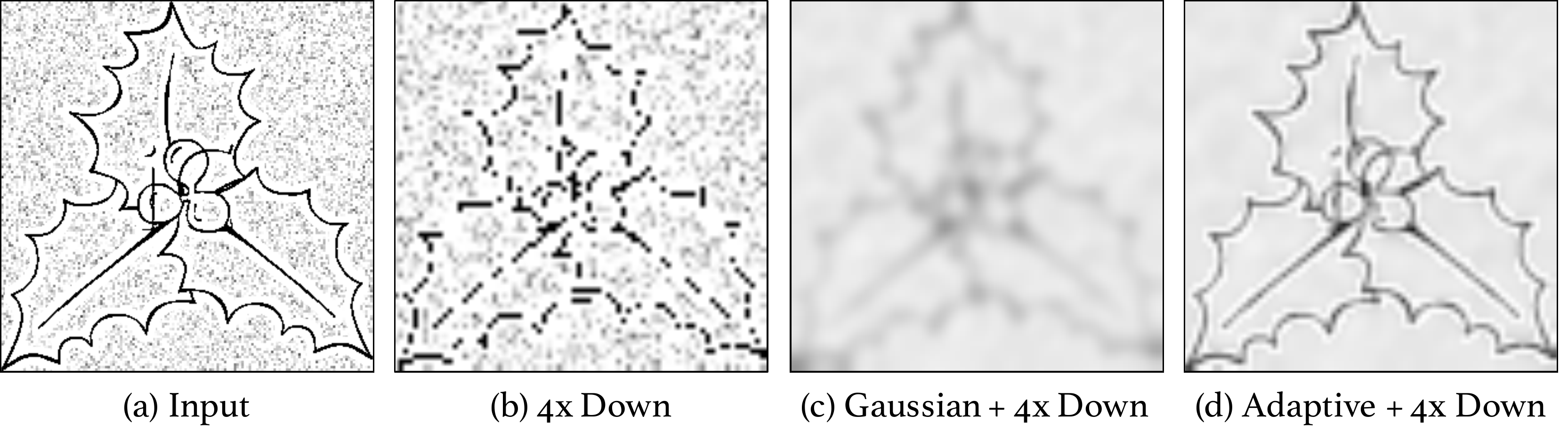}
\end{center}
   \vspace{-0.6cm}
   \caption{\textbf{Effect of adaptive filtering for anti-aliasing.} (a) Input image. (b) Result of direct downsampling. (c) Result of downsampling after applying a single Gaussian filter tuned to match the frequency of the noise. (d) Result of downsampling after applying spatially-adaptive Gaussian filters (stronger blurring for background noise and weaker for edges).}
   \vspace{-20pt}
\label{fig:concept}
\end{figure*}

To tackle this, \cite{zhang2019making} proposed to insert a Gaussian blur layer before each downsampling module in ConvNets. Though simple and effective to a certain degree, we argue that the design choice of applying a universal Gaussian filter is not optimal -- as signal frequencies in a natural image (or feature map) generally vary throughout spatial locations and channels, different blurring filters are needed in order to satisfy the Nyquist theorem to avoid aliasing. 
For example, the image in Fig.~\ref{fig:concept} (a) contains high frequency impulse noise in the background and relatively lower frequency edges in the foreground. Directly applying a downsampling operation produces discontinuous edges and distorted impulse noise shown in (b) due to aliasing. By applying a Gaussian filter before downsampling, we can avoid aliasing as shown in (c). However, as the high frequency impulse noise needs to be blurred more compared to the lower frequency edges, when using a single Gaussian filter tuned for the impulse noise, the edges are over-blurred leading to significant information loss. To solve this issue, what we need is to apply different Gaussian filters to the foreground and background separately, so that we can avoid aliasing while preserving useful information, as in (d). 






With the above observation, we propose a \emph{content-aware anti-aliasing} module, which adaptively predicts low-pass filter weights for different spatial locations. Furthermore, as different feature channels can also have different frequencies (e.g., certain channels capture edges, others capture color blobs), we also predict different filters for different channels. In this way, our proposed module adaptively blurs the input content to avoid aliasing while preserving useful information for downstream tasks. To summarize, our contributions are:


\vspace{-5pt}
\begin{itemize}[leftmargin=*]
    \setlength\itemsep{0em}
	\item We propose a novel adaptive and architecture-independent low-pass filtering layer in ConvNets for anti-aliasing.
	\item We propose novel evaluation metrics, which measure shift consistency for semantic and instance segmentation tasks; i.e., a method's robustness to aliasing effects caused by shifts in the input.
    \item We conduct experiments on image classification (ImageNet), semantic segmentation (PASCAL VOC and Cityscapes), instance segmentation (MS-COCO), and domain generalization (ImageNet to ImageNet VID). The results show that our method outperforms competitive baselines with a good margin on both accuracy and shift consistency.
    \item We demonstrate intuitive qualitative results, which show the interpretability of our module when applied to different spatial locations and channel groups.
\end{itemize}


\vspace{-16pt}
\section{Related Work}

\paragraph{Network robustness}
In deep learning, the robustness problem related to adversarial attacks \cite{szegedy2013intriguing, kurakin2016adversarial}, input translation \cite{mairal2014convolutional, bietti2017invariance, cvpr19uel}, and natural perturbations \cite{shankar2019systematic} has been widely studied. The crux of these studies is how small variations in the input image can lead to large variations in the predictions. In order to obtain a stable and robust network, \cite{xie2019feature, kannan2018adversarial, liao2018defense} introduce novel losses or network architectures to defend against adversarial attacks. Apart from adversarial defense, \cite{mairal2014convolutional, bietti2017invariance} propose new algorithms to learn more shift-invariant representations. Finally, \cite{zhang2019making} provides a new perspective on obtaining shift-invariant features in the context of anti-aliasing. Unlike \cite{zhang2019making}, which applies a single hand-coded low-pass filter regardless of content, we adaptively learn the low-pass filter in a content-aware way and demonstrate it leads to improvement in both recognition accuracy and network robustness.

\vspace{-5pt}
\paragraph{Image filtering}
Low-pass filters like box \cite{rosenberg1974box} and Gaussian \cite{gonzales2002digital} are classic \emph{content agnostic} smoothing filters; i.e., their filter weights are fixed regardless of spatial location and image content. Bilateral \cite{paris2009bilateral} and guided \cite{he2010guided} filters are \emph{content aware} as they can simultaneously preserve edge information while removing noise. Recent works integrate such classic filters into deep networks \cite{hu2017deep, zhang2019making, xie2019feature}. However, directly integrating these modules into a neural network requires careful tuning of hyperparameters subject to the input image (e.g., $\sigma_s$ and $\sigma_r$ in bilateral filter or $r$ and $\epsilon$ in guided filter).  \cite{su2019pixel, jia2016dynamic} introduced the dynamic filtering layer, whose weights are predicted by convolution layers conditioned on pre-computed feature maps. We differ from them in two key aspects: 1) our filter weights vary across both spatial and channel groups, and 2) we insert our low-pass filtering layer before every downsampling layer for anti-aliasing, whereas the dynamic filtering layer is directly linked to the prediction (last) layer in order to incorporate motion information for video recognition tasks. Finally, \cite{wang2019carafe} introduces an adaptive convolution layer for upsampling, whereas we focus on downsampling with an adaptive low-pass filtering layer.

\vspace{-5pt}
\paragraph{Pixel classification tasks} such as semantic segmentation \cite{long2015fully, chen2018encoder} and instance segmentation \cite{he2017mask, bolya2019yolact} require precise modeling of object boundaries, so that pixels from the same object instance can be correctly grouped together.  Thus, while blurring can help reduce aliasing, it can also be harmful to these tasks (e.g., when the edges are blurred too much or not blurred enough hence resulting in aliasing). We investigate the effect of anti-aliasing in these pixel-level tasks, whereas our closest work, \cite{zhang2019making}, focused mainly on image classification.


\vspace{-5pt}
\section{Approach}

\begin{figure*}[t]
\begin{center}
\includegraphics[width=.95\textwidth]{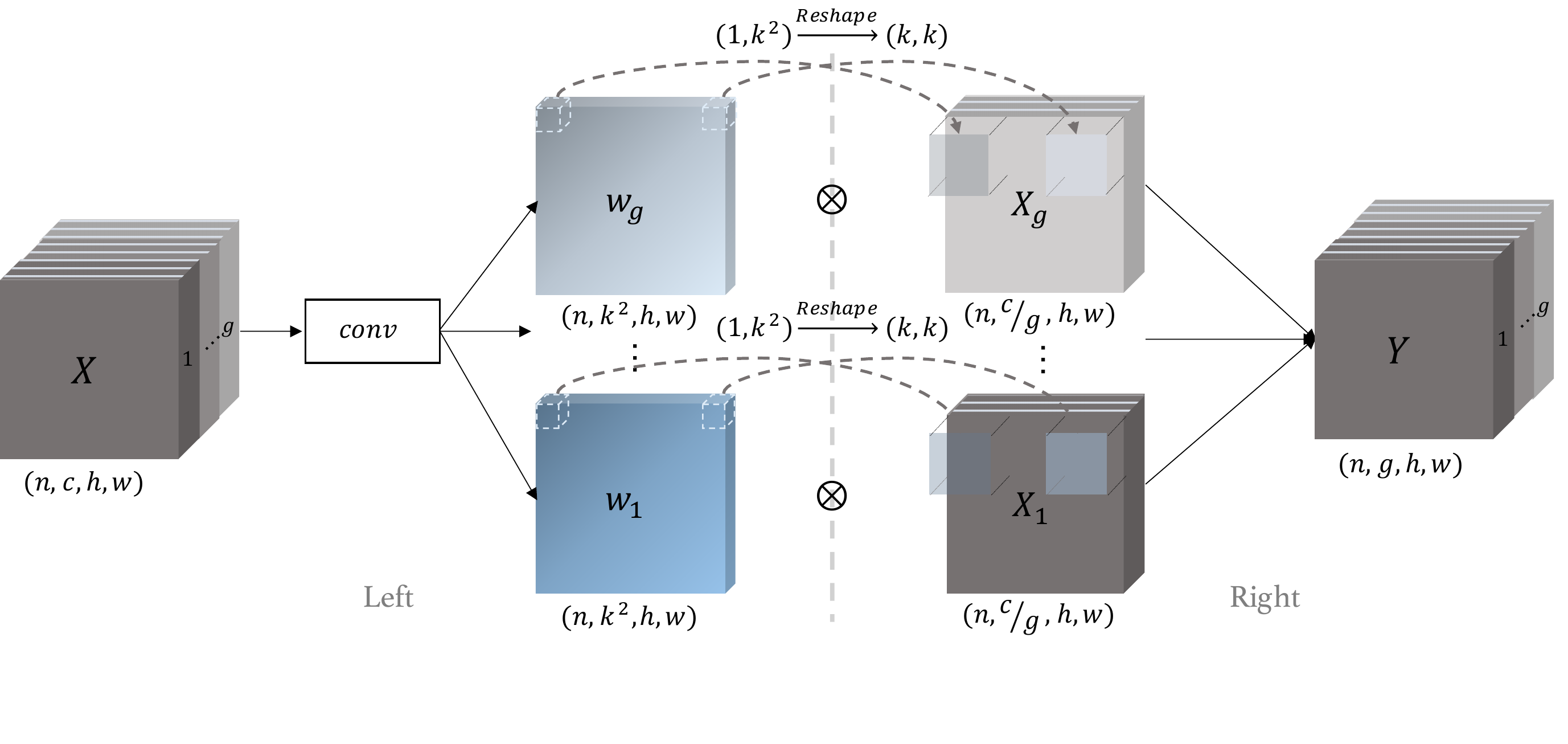}
\end{center}
\vspace{-34pt}
   \caption{\textbf{Method overview.}  (Left) For each spatial location and feature channel group in the input $X$, we predict a $k \times k$ filter $w$.   (Right) We apply the learned filters on $X$ to obtain content aware anti-aliased features. See text for more details.}
\label{fig:method}
\end{figure*}


To enable anti-aliasing for ConvNets, we apply the proposed \emph{content-aware anti-aliasing} module before each downsampling operation in the network. Inside the module, we first generate low-pass filters for \emph{different spatial locations and channel groups} (Fig.~\ref{fig:method} left), and then apply the predicted filters back onto the input features for anti-aliasing (Fig.~\ref{fig:method} right).

\vspace{-5pt}
\paragraph{Spatial adaptive anti-aliasing.} As frequency components can vary across different spatial locations in an image, we propose to learn different low-pass filters in a content-aware manner across spatial locations.  Specifically, given an input feature $X$ that needs to be down-sampled, we generate a low-pass filter $w_{i,j}$ (e.g., a $3 \times 3$ conv filter) for each spatial location $(i,j)$ on $x$. With the predicted low-pass filter $w_{i,j}$, we can then apply it to input $X$:
\begin{equation}
Y_{i,j} =  \sum_{p,q \in \Omega} ~ w_{i,j}^{p,q}~\cdot~X_{i+p,j+q},
\label{eq:spatialadaptive}
\end{equation}
where $Y_{i,j}$ denotes output features at location $(i,j)$ and $\Omega$ points to the set of locations surrounding $(i,j)$ on which we apply the predicted smooth filter.  In this way, the network can learn to blur higher frequency content more than lower frequency content, to reduce undesirable aliasing effects while preserving important content as much as possible.  

\begin{figure*}[t]
\begin{center}
\includegraphics[width=1.\textwidth]{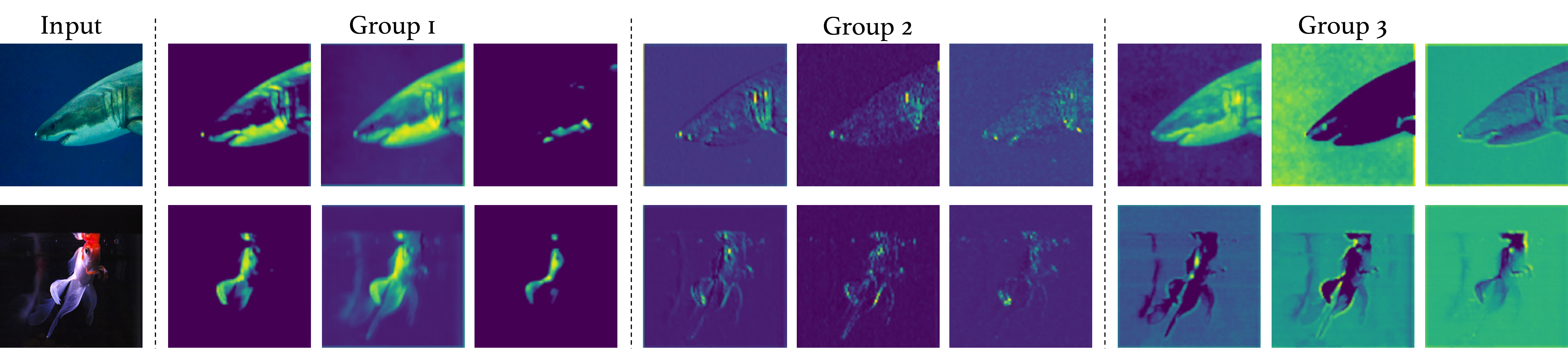}
\end{center}
	\vspace{-13pt}
   \caption{\textbf{Visualization of predicted feature maps within and across groups.} The features within each group are more similar to each other than to those in other groups.  Each group captures a different aspect of the image (e.g., edges, color blobs).}
   \vspace{-3pt}
\label{fig:group}
\end{figure*}

\vspace{-5pt}
\paragraph{Channel-grouped adaptive anti-aliasing.}
Different channels of a feature map can capture different aspects of the input that vary in frequency (e.g., edges, color blobs). Therefore, in addition to predicting different filters for each spatial location, it can also be desirable to predict different filters for each \emph{feature channel}. However, naively predicting a low-pass filter for each spatial location and channel can be computationally very expensive. Motivated by the observation that some channels will capture similar information~\cite{wu-eccv2018}, we group the channels into $k$ groups and predict a single low-pass filter $w_{i,j,g}$ for each group $g$.  Then, we apply $w_{i,j,g}$ to the input $X$:
\begin{equation} 
Y_{i,j}^g =  \sum_{p,q \in \Omega} ~ w_{i,j,g}^{p,q}~\cdot~X^c_{i+p,j+q},
\label{eq:spatialchannelgroupadaptive}
\end{equation}
where $g$ is the group index to which channel $c$ belongs. In this way, channels within a group are learned to be similar, as shown in Fig.~\ref{fig:group}.


%
%

\vspace{-5pt}
\paragraph{Learning to predict filters.} 
To dynamically generate low-pass filters for each spatial location and feature channel group, we apply a convolutional block (conv + batchnorm) to the input feature $X \in R^{n \times c \times h \times w}$ to output $w \in R^{n \times g \times k^2 \times h \times w}$, where $g$ denotes the number of channel groups and each of the $k^2$ channels corresponds to an element in one of $k \times k$ locations in the filters. For grouping, we group every $c/g$ consecutive channels, where $c$ is the total number of channels.  Finally, to ensure that the generated filters are low-pass, we constrain their weights to be positive and sum to one by passing it through a softmax layer.


\begin{figure*}[t]
\begin{center}
\includegraphics[width=1.\textwidth]{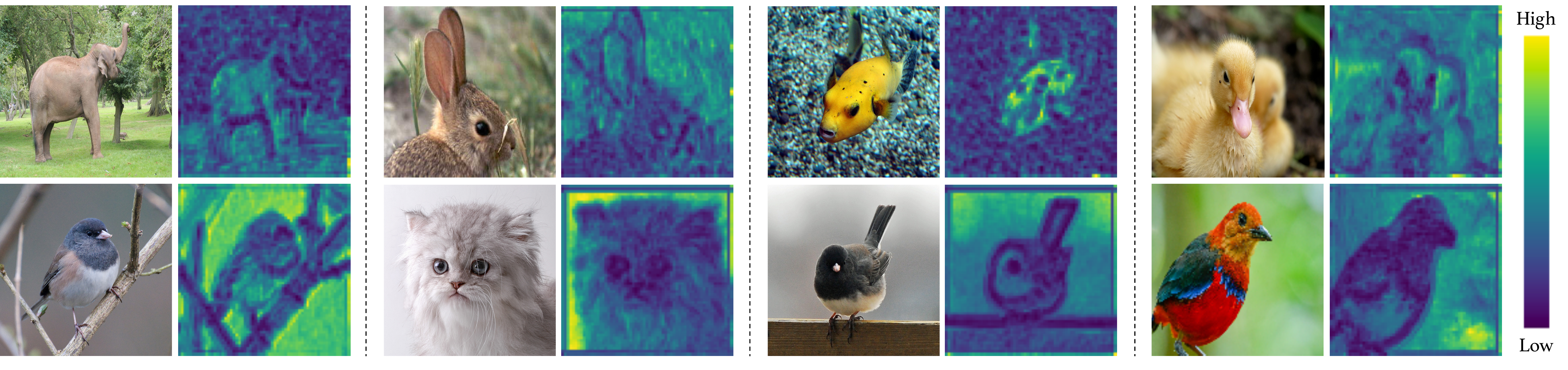}
\end{center}
   \vspace{-14pt}
   \caption{\textbf{Variance of the learned filter weights across spatial locations.} Low variance corresponds to more blur, while high variance corresponds to less blur. Our model correctly learns to blur high frequency content (e.g., edges) more to prevent aliasing, and blur low frequency content less to preserve useful information.}
   \vspace{-4pt}
\label{fig:spatial}
\end{figure*}

\vspace{-5pt}
\paragraph{Analyzing the predicted filters.}

In this section, we analyze the behavior of our learned filters.  First, we analyze how the filters spatially adapt to different image content.  For this, we compute the variance of the learned filter weights across different spatial locations.  A $k \times k$ average filter with $1/k^2$ intensity in each element will have zero variance whereas an identity filter with one in the center and zeros everywhere else will have high variance. From Fig.~\ref{fig:spatial}, one can clearly see that when the image content has high frequency information (e.g., elephant background trees, bird contours), the learned filters' variance tends to be smaller; i.e., more blur is needed to prevent aliasing.  Conversely, the filters' variance is larger when the content is relatively smoother (e.g., background in bird images); i.e., less blur is needed to prevent aliasing.  In this way, the learned filters can reduce aliasing during sampling while preserving useful image content as much as possible.

We next analyze how the filters adapt to different content across different feature groups.  Fig.~\ref{fig:group} shows this effect; e.g., group 1 captures relatively low frequency information with smooth areas, while group 2 captures higher frequency information with sharp intensity transitions. In this way, the learned filters can adapt to different frequencies across feature channels, while saving computational costs by learning the same filter per group.



\vspace{-3pt}
\section{Experiments}
\vspace{-3pt}

We first introduce our experimental settings and propose consistency metrics for image classification, instance segmentation, and semantic segmentation. We compare to strong baselines including ResNet \cite{he2016deep}, Deeplab v3+ \cite{chen2018encoder}, Mask R-CNN on large scale datasets including ImageNet, ImageNet VID \cite{deng2009imagenet}, MS COCO \cite{lin2014microsoft}, PASCAL VOC \cite{everingham2015pascal} and Cityscapes \cite{cordts2016cityscapes}. We also conduct ablation studies on our design choices including number of groups, parameter counts, as well as filter types. Finally, we present qualitative results demonstrating the interpretability of our anti-aliasing module.

\begin{table} 
\centering
\footnotesize
\begin{tabular}{c|c|cc|cc|cc} 
\multicolumn{1}{c}{}   & \multicolumn{1}{c}{} & \multicolumn{2}{c}{accuracy} & \multicolumn{2}{c}{consistency } & \multicolumn{2}{l}{generalization}  
        \\
methods      &   Filter     & Abs             & Delta                & Abs            & Delta                   & Abs  & Delta                                \\ 
\hline
ResNet-101 \cite{he2016deep} & -      & 77.7            & -                    & 90.6           & -                       & 67.6 & -                                    \\ 
\hline
\multirow{2}{*}{LPF \cite{zhang2019making}}       & 3 x 3  & 78.4            & + 0.7                & 91.6           & + 1.0                   & 68.8 & +1.2                                 \\
           & 5 x 5  & 77.7            & + 0.0                & 91.8           & + 1.2                   & 67.0    & - 0.6                                    \\ 
\hline
\multirow{2}{*}{Ours}       & 3 x 3  & \textbf{79.0 }  & \textbf{+ 1.3 }      & 91.8           & + 1.2                   & \textbf{69.9} & \textbf{+2.3}                                 \\
           & 5 x 5  & 78.6            & + 0.9                & \textbf{92.2}  & \textbf{+ 1.6}          & 69.1    & +1.5                                    \\
\end{tabular}
\vspace{6pt}
\caption{Image classification accuracy, consistency on ImageNet \cite{deng2009imagenet}, and domain generalization results ImageNet $\rightarrow$ ImageNet VID \cite{deng2009imagenet}. We compare to strong ResNet-101 \cite{he2016deep} and LPF (low-pass filter) \cite{zhang2019making} baselines. Our method shows consistent improvement in accuracy, consistency, and generalization.}
\label{table:imagenet}
\vspace{-5pt}
\end{table}

\vspace{-3pt}
\subsection{Image Classification}\label{sec:imageclassification}
\vspace{-3pt}

\paragraph{Experimental settings}
We evaluate on ILSVRC2012~\cite{deng2009imagenet}, which contains 1.2M training and 50K validation images for 1000 object classes. We use input image size of $224 \times 224$, SGD solver with initial learning rate 0.1, momentum 0.9, and weight decay 1e-4. Full training schedule is 90 epochs with 5 epoch linear scaling warm up. Learning rate is reduced by 10x every 30 epochs. We train on 4 GPUs, with batch size 128 and batch accumulation of 2. For fair comparison, we use the same set of hyperparameters and training schedule for both ResNet-101, LPF~\cite{zhang2019making} baselines as well as our method. The number of groups is set to 8 according to our ablation study. We extend the code base introduced in \cite{zhang2019making}. 

\vspace{-5pt}
\paragraph{Consistency metric}
We use the consistency metric defined in \cite{zhang2019making}, which measures how often the model outputs the same top-1 class given two different shifts on the same test image: $Consistency = \mathbb{E}_{X,h_1, w_1, h_2, w_2}~\mathbb{I} \{ F(X_{h_1,w_1}) = F(X_{h_2,w_2})\}$, where $\mathbb{E}$ and $\mathbb{I}$ denote expectation and indicator function (outputs 1/0 with true/false inputs). $X$ is the input image, $h_1, w_1$ (height/width) and $ h_2, w_2$ parameterize the shifts and $F(\cdot)$ denotes the predicted top-1 class. 


\vspace{-5pt}
\paragraph{Results and analysis}
As shown in Table \ref{table:imagenet}, our adaptive anti-aliasing module outperforms the baseline ResNet-101 without anti-aliasing with a $1.3$ point boost (79.0 vs 77.7) in top-1 accuracy on ImageNet classification.  More importantly, when comparing to LPF~\cite{zhang2019making}, which uses a fixed blurring kernel for anti-aliasing, our method scores $0.6$ points higher (79.0 vs 78.4) on top-1 accuracy.  Furthermore, our method not only achieves better classification accuracy, it also outputs more consistent results (+0.2/+0.4 consistency score improvements for 3$\times$3 and 5$\times$5 filter sizes) compared to LPF. These results reveal that our method preserves more discriminative information for recognition when blurring feature maps.


\vspace{-9pt}
\subsection{Domain Generalization}

\paragraph{Experimental settings}
ImageNet VID is a video object detection dataset, which has 30 classes that overlap with $284$ classes in ImageNet (some classes in ImageNet VID are the super class of ImageNet). It contains $3862/1315$ training/validation videos. We randomly select three frames from each validation video, and evaluate Top-1 accuracy on them to measure the generalization capability of our model which is pretrained on ImageNet (i.e. it has never seen any frame in ImageNet VID). As a video frame may contain multiple objects in different classes, we count a prediction as correct as long as it belongs to one of the ground-truth classes.

\vspace{-5pt}
\paragraph{Results and analysis}
Table \ref{table:imagenet} reveals that our method generalizes better to a different domain compared to the ResNet-101 baseline (+$2.3$\% points increase in top-1 accuracy for $3 \times 3$ filter) and LPF model (+$1.1$\%) which adopts a fixed blur kernel. We hypothesize that the better generalization capability comes from the fact that we learn a representation that is less sensitive to downsampling (i.e., more robust to shifts). This is particularly useful for video frames, as they can be thought of as having natural shift perturbations of the same content across frames~\cite{shankar2019systematic}.





\vspace{-6pt}
\subsection{Instance Segmentation}

\paragraph{Experimental settings}
In this section, we present results on MS-COCO for instance segmentation~\cite{lin2014microsoft}. MS-COCO contains 330k images, 1.5M object instances and 80 categories. We use Mask R-CNN \cite{he2017mask} as our base architecture. We adopt the hyperparameter settings from the implementation of~\cite{massa2018mrcnn}. When measuring consistency, we first resize images to $800 \times 800$ and then take a crop of $736 \times 736$ as input.

\begin{figure}[t]
\begin{center}
\includegraphics[width=.9\textwidth]{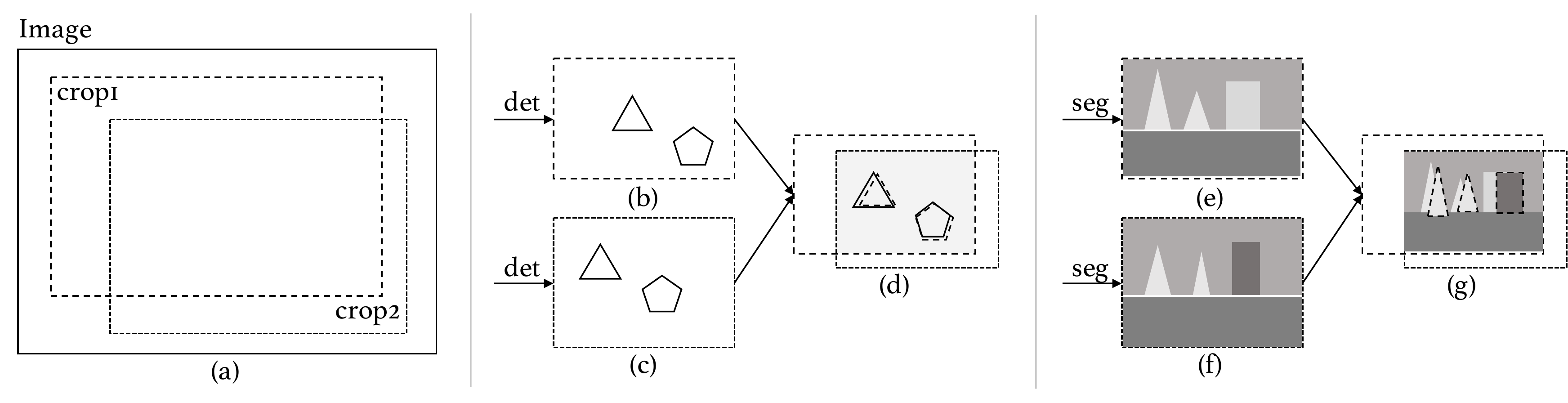}
\end{center}
\vspace{-18pt}
   \caption{Our new consistency metrics. (b,c,d): mean Average Instance Segmentation Consistency (mAISC). (e,f,g): mean Average Semantic Segmentation Consistency (mASSC).}
\label{fig:consist}
\end{figure}

\vspace{-5pt}
\paragraph{Consistency metric (mAISC)}
We propose a new mean Average Instance Segmentation Consistency (mAISC) metric to measure the shift invariance property of instance segmentation methods. As shown in Fig.~\ref{fig:consist}, given an input image (a), we randomly select two crops (b) and (c), and apply an instance segmentation method on them separately. $M(b)$ and $M(c)$ denote the predicted instances in the overlapping region of image (b) and (c). To measure consistency, for any given instance $m_b$ in $M(b)$ we find its highest overlapping counterpart $m_c$ in $M(c)$. If the IOU between $m_b$ and $m_c$ is larger than a threshold (0.9 in our experiments), we regard $m_b$ as a positive (consistent) sample in $M(b)$. (A sample $m_c$ from $M(c)$ can only be considered a counterpart of any instance in $M(b)$ once.) We compute the final mAISC score as the mean percentage of positive samples in $M(b)$ over all input image pairs.

\vspace{-5pt}
\paragraph{Results and analysis}
We evaluate mAP and mAISC for both mask and box predictions.  As shown in Table~\ref{table:inst-seg}, while simply applying a fixed Gaussian low-pass filter improves mAP by +0.7/+0.8 points for mask/box, our adaptive content-aware anti-aliasing module is more effective (further +0.4/+0.5 point improvement over LPF for mask/box). This demonstrates that it is important to have different low-pass filters for different spatial locations and channel groups. More interestingly, by introducing our adaptive low-pass filters, mAISC increases by a large margin (+5.1/+4.7 for mask/box over the baseline, and +1.0/+1.0 over LPF). This result demonstrates that 1) an anti-aliasing module significantly improves shift consistency via feature blurring, and 2) edges (higher frequency) are better preserved using our method (compared to LPF) during downsampling which are critical for pixel classification tasks.

\vspace{-5pt}
\subsection{Semantic Segmentation}

\paragraph{Experimental settings}
We next evaluate on PASCAL VOC2012 \cite{everingham2015pascal} and Cityscapes \cite{cordts2016cityscapes} semantic segmentation with Deeplab v3+ \cite{chen2018encoder} as the base model. We extend implementations from \cite{hu2020temporally} and \cite{VainF_2019}. For Cityscapes, we use syncBN with a batch size of 8. As for PASCAL VOC, we use a batch size of 16 on two GPUs without syncBN. We report better performance compared to the original implementation for DeepLab v3+ on PASCAL VOC. For Cityscapes, our ResNet-101 backbone outperforms the Inception backbone used in \cite{chen-deeplab2017}.

\begin{table*}[t]

\centering
\footnotesize
\begin{tabular}{l|cccc|cccc} 
\multicolumn{1}{c}{}    & \multicolumn{4}{c}{Mask}                               & \multicolumn{4}{c}{Box}                                  \\
\hline
method     & mAP            & Delta           & mAISC           & Delta          & mAP            & Delta           & mAISC           & Delta           \\ 
\hline
Mask R-CNN \cite{he2017mask} & 36.1           & -               & 62.9          & -              & 40.1           & -               & 65.1          & -               \\
LPF \cite{zhang2019making}        & 36.8           & + 0.7           & 66.0          & + 4.1          & 40.9           & + 0.8           & 68.8          & + 3.7~          \\
Ours       & \textbf{37.2}  & \textbf{+ 1.1}  & \textbf{67.0} & \textbf{+ 5.1} & \textbf{41.4}  & \textbf{+ 1.3}  & \textbf{69.8} & \textbf{+ 4.7}  \\
\end{tabular}
\vspace{2pt}
\caption{Instance segmentation results on MS COCO. We compare to Mask R-CNN~\cite{he2017mask} and LPF~\cite{zhang2019making}.  Our approach consistently improves over the baselines for both mask and box accuracy and consistency.}
\label{table:inst-seg}
\end{table*}
%



\begin{table}[t]
\centering
\footnotesize
\begin{tabular}{l|cccc|cccc} 
\multicolumn{1}{l}{} & \multicolumn{4}{c}{PASCAL VOC}                                       & \multicolumn{4}{c}{Cityscapes}                                          \\ 
\hline
method                & mIOU           & Delta           & mASSC                  & Delta          & mIOU          & Delta          & mASSC                  & Delta           \\ 
\hline
Deeplab v3+ \cite{chen2018encoder}            & 78.5           & -               & 95.5$\pm$0.11          & -              & 78.5          & -              & 96.0$\pm0.10$          & -               \\
LPF \cite{zhang2019making}                  & 79.4           & + 0.9           & 95.9$\pm$0.07          & + 0.4          & 78.9          & + 0.4          & 96.1$\pm$0.05          & + 0.1           \\
Ours                  & \textbf{80.3}  & \textbf{+ 1.8}  & \textbf{96.0$\pm$0.13} & \textbf{+ 0.5} & \textbf{79.5} & \textbf{+ 1.0} & \textbf{96.3$\pm$0.07} & \textbf{+ 0.3}  \\
\end{tabular}
\vspace{2pt}
\caption{Semantic segmentation on PASCAL VOC 2012 \cite{everingham2015pascal} and Cityscapes \cite{cordts2016cityscapes}. We compare to Deeplab v3+~\cite{chen2018encoder} and LPF~\cite{zhang2019making}. Our approach leads to improved accuracy and consistency.}
\label{table:semanticseg}
\end{table}





\vspace{-5pt}
\paragraph{Consistency metric (mASSC)} We propose a new
mean Average Semantic Segmentation Consistency (mASSC) metric to measure shift consistency for semantic segmentation methods. Similar to mAISC, we take two random crops (e,f) from the input image (a) in Fig. \ref{fig:consist}. We then compute the Semantic Segmentation Consistency between the overlapping regions $X$ and $Y$ of the two crops:  $Consistency(X,Y) := \mathbb{E}_{i\in[0,h)}\mathbb{E}_{j\in[0,w)}~\mathbb{I}[S(X)_{i,j} = S(Y)_{i,j}]$, where $S(X)_{i,j}$ and $S(Y)_{i,j}$ denote the predicted class label of pixel $(i,j)$ in $X$ and $Y$, and $h,w$ is the height and width of the overlapping region. We average this score for all pairs of crops in an image, and average those scores over all test images to compute the final mASSC.



\vspace{-5pt}
\paragraph{Results and analysis}
As shown in Table~\ref{table:semanticseg}, our method improves mIOU by 1.8 and 1.0 points on PASCAL VOC and Cityscapes compared to the strong baseline of DeepLab v3+. 
Furthermore, our method also consistently improves the mASSC score (+0.5 and +0.3 for VOC and Cityscapes) despite the high numbers achieved by the baseline method (95.5/96.0).
Finally, to measure the variance of our mASSC results, we report the standard deviation over three runs with different random seeds.

\vspace{-6pt}
\subsection{Ablation Studies}

\paragraph{Experimental settings} For efficiency, we perform all ablation studies using ResNet-18 with input image size $112 \times 112$ and batch size 200 on ImageNet. All other hyperparameters are identical to those used in Sec.~\ref{sec:imageclassification}.

\begin{figure}[t]
\centerline{\includegraphics[width=.83\textwidth]{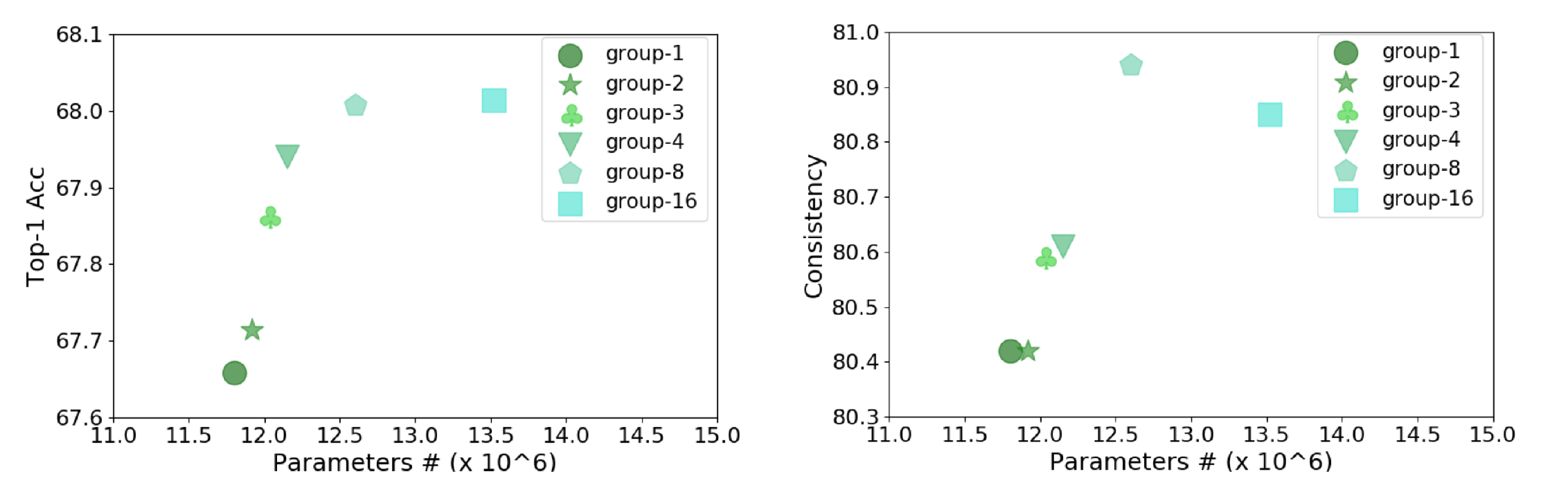}}
\vspace{-9pt}
\caption{Effect of number of groups on top-1 accuracy and consistency.}
\label{fig:aba_group}
\end{figure}

\begin{SCtable}[]
 \footnotesize
	\begin{tabular}{l|cc} 
    methods & top-1 Acc & consistency           \\ 
    \hline
    ResNet         & 66.5 & 79.1  \\
    Gaussian     & 66.7                     & 79.8                  \\
    Image Adaptive   & 66.7                     & 78.7                  \\
    Spatial Adaptive & 67.7                     & 80.3                  \\
    Ours & \textbf{68.0}                     & \textbf{80.9}                  \\
    \end{tabular}
	\caption{Filter ablations. Gaussian blur is better than no blur (ResNet). Learning the blur filter globally (Image Ada.), spatially (Spatial Ada.), and over channels (Ours) progressively does better.}
	\label{table:ablation}
\end{SCtable}

\vspace{-6pt}
\paragraph{Number of channel groups.}
We vary the number of channel groups and study its influence on image classification accuracy. As shown in Fig.~\ref{fig:aba_group}, the trend is clear -- increasing the number of groups generally leads to improved top-1 accuracy. This demonstrates the effectiveness of predicting different filters across channels. However, there exists a diminishing return in this trend -- the performance saturates when the group number goes beyond 8. We hypothesize this is caused by overfitting.


\vspace{-7pt}
\paragraph{Number of parameters.}
We further compare the effects of directly increasing the number of parameters in the base network \emph{vs} adding more groups in our content-aware low-pass filters. To increase the number of parameters for the base network, we increase the base channel size in ResNet-18. We find that directly increasing the number of parameters barely improves top-1 accuracy -- when the number of parameters increases from 12.17M to 12.90M, top-1 accuracy increases only by 0.1\%. Also, with similar (or less) number of parameters, our method yields a higher performance gain compared to naively increasing network capacity (68.0\% \emph{vs} 67.7\% top-1 accuracy for 12.60M \emph{vs} 12.90M parameters). This shows that our adaptive anti-aliasing method does not gain performance by simply scaling up its capacity.  

\vspace{-6pt}
\paragraph{Type of filter.}
In Table \ref{table:ablation}, we ablate our pixel adaptive filtering layers with various baseline components. Applying the same low-pass filter (Gaussian, Image Adaptive) across the entire image performs better than the vanilla ResNet-18 without any anti-aliasing. Here, Image Adaptive refers to the baseline which predicts a single low-pass filter for the entire image.  By adaptively learning a spatially variant low-pass filter, performance improves further (Spatial Adaptive).  Overall, our method achieves the best performance which demonstrates the benefits of predicting filters that are both spatially varying and channel adaptive. 

\vspace{-6pt}
\paragraph{Overhead.}
Finally, with our spatial/channel adaptive filtering added, the number of parameters increases by 2.9-7.8\% for ResNet models (e.g., 4\% for R-101, 4.5M to 4.63M). As for runtime, on a RTX2070 GPU, our method (R-101 backbone) takes 6.4 ms to forward a 224x224 image whereas a standard ResNet-101 takes 4.3 ms.

\vspace{-3pt}
\subsection{Qualitative Results}

\begin{figure*}[t]
\begin{center}
\includegraphics[width=1.\textwidth]{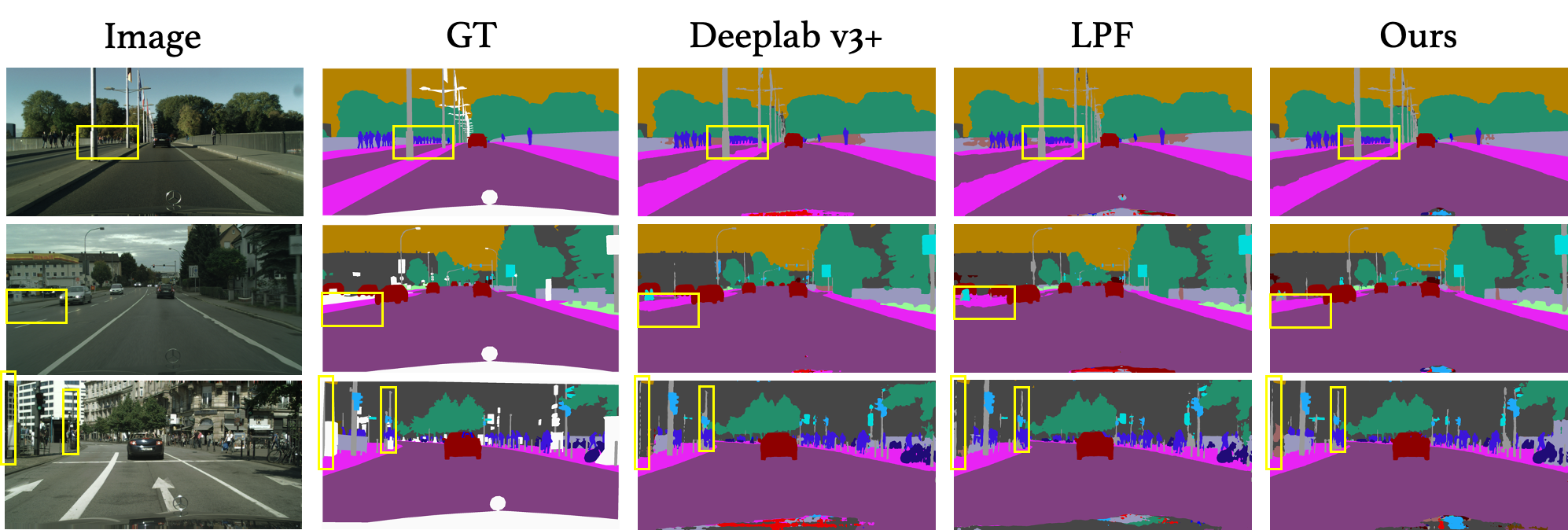}
\end{center}
   \vspace{-14pt}
   \caption{Qualitative results for semantic segmentation on Cityscapes.}
   \vspace{-3pt}
\label{fig:sup4}
\end{figure*}

\paragraph{Semantic Segmentation.}
We show qualitative results for semantic segmentation in Fig.~\ref{fig:sup4} to demonstrate that our module better preserves edge information. For example, in the first row, within the yellow box region, our method clearly distinguishes the road edge compared to Deeplab v3+ and LPF. Similar behavior (better segmented road contours) is also observed in the second row. This holds for other objects as well -- the light pole has better delineation compared to both baselines in the third row.

\begin{figure}[t]
\centerline{\includegraphics[width=.99\textwidth]{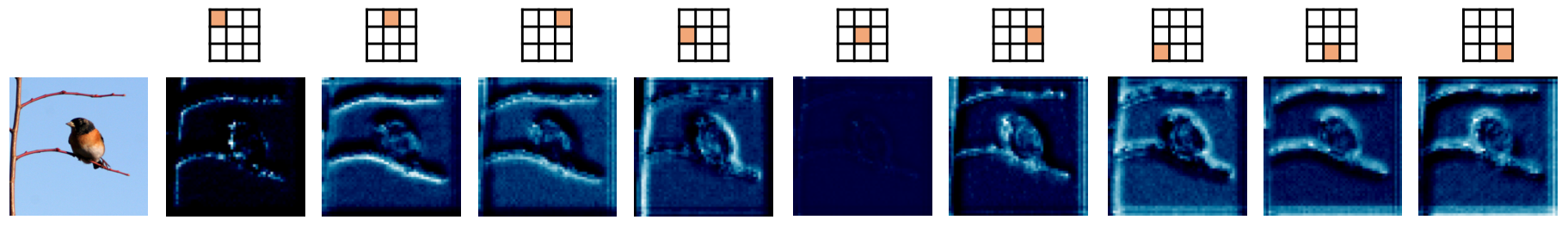}}
\vspace{-8pt}
\caption{Visualization of learned filter weights at each spatial location.}
\vspace{-6pt}
\label{fig:qualitative}
\end{figure}

\vspace{-5pt}
\paragraph{Low-pass filter weights.}
To further understand our adaptive filtering module, we visualize the low-pass filter weights for each spatial location. As shown in Fig.~\ref{fig:qualitative}, our model tends to ``grow'' edges so that it's easier for them to be preserved. For example, the learned filter tends to integrate more information from left to right (see center-left and bottom-left weights in Fig.~\ref{fig:qualitative}) on the vertical tree branch and thus grow it to be thicker. This way, it's easier for tree branch contours to be preserved after downsampling.

\vspace{-5pt}
\section{Conclusion}
\vspace{-5pt}
In this paper, we proposed an adaptive content-aware low-pass filtering layer, which predicts separate filter weights for each spatial location and channel group of the input. We quantitatively demonstrated the effectiveness of the proposed method across multiple tasks and qualitatively showed that our approach effectively adapts to the different feature frequencies to avoid aliasing while preserving useful information for recognition.

\section{Acknowledgements}
This work was supported in part by ARO YIP W911NF17-1-0410, NSF CAREER IIS-1751206, NSF CCF-1934568, GCP research credit program, and AWS ML research award.

\bibliography{egbib}

\begin{thebibliography}{37}
\providecommand{\natexlab}[1]{#1}
\providecommand{\url}[1]{\texttt{#1}}
\expandafter\ifx\csname urlstyle\endcsname\relax
  \providecommand{\doi}[1]{doi: #1}\else
  \providecommand{\doi}{doi: \begingroup \urlstyle{rm}\Url}\fi

\bibitem[Azulay and Weiss(2018)]{azulay2018deep}
Aharon Azulay and Yair Weiss.
\newblock Why do deep convolutional networks generalize so poorly to small
  image transformations?
\newblock In \emph{JMLR}, 2018.

\bibitem[Bietti and Mairal(2017)]{bietti2017invariance}
Alberto Bietti and Julien Mairal.
\newblock Invariance and stability of deep convolutional representations.
\newblock In \emph{NeurIPS}, 2017.

\bibitem[Bolya et~al.(2019)Bolya, Zhou, Xiao, and Lee]{bolya2019yolact}
Daniel Bolya, Chong Zhou, Fanyi Xiao, and Yong~Jae Lee.
\newblock {YOLACT: real-time instance segmentation}.
\newblock In \emph{ICCV}, 2019.

\bibitem[Chen et~al.(2017)Chen, Papandreou, Schroff, and
  Adam]{chen-deeplab2017}
Liang-Chieh Chen, George Papandreou, Florian Schroff, and Hartwig Adam.
\newblock Rethinking atrous convolution for semantic image segmentation.
\newblock \emph{arXiv preprint arXiv:1706.05587}, 2017.

\bibitem[Chen et~al.(2018)Chen, Zhu, Papandreou, Schroff, and
  Adam]{chen2018encoder}
Liang-Chieh Chen, Yukun Zhu, George Papandreou, Florian Schroff, and Hartwig
  Adam.
\newblock Encoder-decoder with atrous separable convolution for semantic image
  segmentation.
\newblock In \emph{ECCV}, 2018.

\bibitem[Cordts et~al.(2016)Cordts, Omran, Ramos, Rehfeld, Enzweiler, Benenson,
  Franke, Roth, and Schiele]{cordts2016cityscapes}
Marius Cordts, Mohamed Omran, Sebastian Ramos, Timo Rehfeld, Markus Enzweiler,
  Rodrigo Benenson, Uwe Franke, Stefan Roth, and Bernt Schiele.
\newblock The cityscapes dataset for semantic urban scene understanding.
\newblock In \emph{CVPR}, 2016.

\bibitem[Deng et~al.(2009)Deng, Dong, Socher, Li, Li, and
  Fei-Fei]{deng2009imagenet}
Jia Deng, Wei Dong, Richard Socher, Li-Jia Li, Kai Li, and Li~Fei-Fei.
\newblock Imagenet: A large-scale hierarchical image database.
\newblock In \emph{CVPR}, 2009.

\bibitem[Everingham et~al.(2015)Everingham, Eslami, Van~Gool, Williams, Winn,
  and Zisserman]{everingham2015pascal}
Mark Everingham, SM~Ali Eslami, Luc Van~Gool, Christopher~KI Williams, John
  Winn, and Andrew Zisserman.
\newblock The pascal visual object classes challenge: A retrospective.
\newblock In \emph{IJCV}, 2015.

\bibitem[Gonzales and Woods(2002)]{gonzales2002digital}
Rafael~C Gonzales and Richard~E Woods.
\newblock Digital image processing, 2002.

\bibitem[He et~al.(2010)He, Sun, and Tang]{he2010guided}
Kaiming He, Jian Sun, and Xiaoou Tang.
\newblock Guided image filtering.
\newblock In \emph{ECCV}, 2010.

\bibitem[He et~al.(2015)He, Zhang, Ren, and Sun]{he-iccv2015}
Kaiming He, Xiangyu Zhang, Shaoqing Ren, and Jian Sun.
\newblock Delving deep into rectifiers: Surpassing human-level performance on
  imagenet classification.
\newblock In \emph{ICCV}, 2015.

\bibitem[He et~al.(2016)He, Zhang, Ren, and Sun]{he2016deep}
Kaiming He, Xiangyu Zhang, Shaoqing Ren, and Jian Sun.
\newblock Deep residual learning for image recognition.
\newblock In \emph{CVPR}, 2016.

\bibitem[He et~al.(2017)He, Gkioxari, Doll{\'a}r, and Girshick]{he2017mask}
Kaiming He, Georgia Gkioxari, Piotr Doll{\'a}r, and Ross Girshick.
\newblock Mask r-cnn.
\newblock In \emph{ICCV}, 2017.

\bibitem[Hu et~al.(2017)Hu, Shuai, Liu, and Wang]{hu2017deep}
Ping Hu, Bing Shuai, Jun Liu, and Gang Wang.
\newblock Deep level sets for salient object detection.
\newblock In \emph{CVPR}, 2017.

\bibitem[Hu et~al.(2020)Hu, Heilbron, Wang, Lin, Sclaroff, and
  Perazzi]{hu2020temporally}
Ping Hu, Fabian~Caba Heilbron, Oliver Wang, Zhe Lin, Stan Sclaroff, and
  Federico Perazzi.
\newblock Temporally distributed networks for fast video semantic segmentation.
\newblock In \emph{CVPR}, 2020.

\bibitem[Jia et~al.(2016)Jia, De~Brabandere, Tuytelaars, and
  Gool]{jia2016dynamic}
Xu~Jia, Bert De~Brabandere, Tinne Tuytelaars, and Luc~V Gool.
\newblock Dynamic filter networks.
\newblock In \emph{NeurIPS}, 2016.

\bibitem[Kannan et~al.(2018)Kannan, Kurakin, and
  Goodfellow]{kannan2018adversarial}
Harini Kannan, Alexey Kurakin, and Ian Goodfellow.
\newblock Adversarial logit pairing.
\newblock \emph{arXiv preprint arXiv:1803.06373}, 2018.

\bibitem[Kurakin et~al.(2016)Kurakin, Goodfellow, and
  Bengio]{kurakin2016adversarial}
Alexey Kurakin, Ian Goodfellow, and Samy Bengio.
\newblock Adversarial examples in the physical world.
\newblock \emph{arXiv preprint arXiv:1607.02533}, 2016.

\bibitem[Liao et~al.(2018)Liao, Liang, Dong, Pang, Hu, and
  Zhu]{liao2018defense}
Fangzhou Liao, Ming Liang, Yinpeng Dong, Tianyu Pang, Xiaolin Hu, and Jun Zhu.
\newblock Defense against adversarial attacks using high-level representation
  guided denoiser.
\newblock In \emph{CVPR}, 2018.

\bibitem[Lin et~al.(2014)Lin, Maire, Belongie, Hays, Perona, Ramanan,
  Doll{\'a}r, and Zitnick]{lin2014microsoft}
Tsung-Yi Lin, Michael Maire, Serge Belongie, James Hays, Pietro Perona, Deva
  Ramanan, Piotr Doll{\'a}r, and C~Lawrence Zitnick.
\newblock Microsoft coco: Common objects in context.
\newblock In \emph{ECCV}, 2014.

\bibitem[Long et~al.(2015)Long, Shelhamer, and Darrell]{long2015fully}
Jonathan Long, Evan Shelhamer, and Trevor Darrell.
\newblock Fully convolutional networks for semantic segmentation.
\newblock In \emph{CVPR}, 2015.

\bibitem[Mairal et~al.(2014)Mairal, Koniusz, Harchaoui, and
  Schmid]{mairal2014convolutional}
Julien Mairal, Piotr Koniusz, Zaid Harchaoui, and Cordelia Schmid.
\newblock Convolutional kernel networks.
\newblock In \emph{NeurIPS}, 2014.

\bibitem[Massa and Girshick(2018)]{massa2018mrcnn}
Francisco Massa and Ross Girshick.
\newblock {maskrcnn-benchmark: Fast, modular reference implementation of
  Instance Segmentation and Object Detection algorithms in PyTorch}.
\newblock \url{https://github.com/facebookresearch/maskrcnn-benchmark}, 2018.
\newblock Accessed: [Oct.10 2019].

\bibitem[Mnih et~al.(2015)Mnih, Kavukcuoglu, Silver, Rusu, Veness, Bellemare,
  Graves, Riedmiller, Fidjeland, Ostrovski, Petersen, Beattie, Sadik,
  Antonoglou, King, Kumaran, Wierstra, Legg, and Hassabis]{mnih-nature2015}
Volodymyr Mnih, Koray Kavukcuoglu, David Silver, Andrei~A. Rusu, Joel Veness,
  Marc~G. Bellemare, Alex Graves, Martin Riedmiller, Andreas~K. Fidjeland,
  Georg Ostrovski, Stig Petersen, Charles Beattie, Amir Sadik, Ioannis
  Antonoglou, Helen King, Dharshan Kumaran, Daan Wierstra, Shane Legg, and
  Demis Hassabis.
\newblock Human-level control through deep reinforcement learning.
\newblock \emph{Nature}, 518\penalty0 (1):\penalty0 529--533, 2015.

\bibitem[Paris et~al.(2009)Paris, Kornprobst, Tumblin, Durand,
  et~al.]{paris2009bilateral}
Sylvain Paris, Pierre Kornprobst, Jack Tumblin, Fr{\'e}do Durand, et~al.
\newblock Bilateral filtering: Theory and applications.
\newblock \emph{Foundations and Trends{\textregistered} in Computer Graphics
  and Vision}, 4\penalty0 (1):\penalty0 1--73, 2009.

\bibitem[Rajpurkar et~al.(2016)Rajpurkar, Zhang, Lopyrev, and
  Liang]{rajpurkar2016squad}
Pranav Rajpurkar, Jian Zhang, Konstantin Lopyrev, and Percy Liang.
\newblock Squad: 100,000+ questions for machine comprehension of text.
\newblock In \emph{EMNLP}, 2016.

\bibitem[Rosenberg(1974)]{rosenberg1974box}
D~Rosenberg.
\newblock Box filter, June~11 1974.
\newblock US Patent 3,815,754.

\bibitem[Shankar et~al.(2019)Shankar, Dave, Roelofs, Ramanan, Recht, and
  Schmidt]{shankar2019systematic}
Vaishaal Shankar, Achal Dave, Rebecca Roelofs, Deva Ramanan, Benjamin Recht,
  and Ludwig Schmidt.
\newblock A systematic framework for natural perturbations from videos.
\newblock \emph{arXiv preprint arXiv:1906.02168}, 2019.

\bibitem[Su et~al.(2019)Su, Jampani, Sun, Gallo, Learned-Miller, and
  Kautz]{su2019pixel}
Hang Su, Varun Jampani, Deqing Sun, Orazio Gallo, Erik Learned-Miller, and Jan
  Kautz.
\newblock Pixel-adaptive convolutional neural networks.
\newblock In \emph{CVPR}, 2019.

\bibitem[Szegedy et~al.(2013)Szegedy, Zaremba, Sutskever, Bruna, Erhan,
  Goodfellow, and Fergus]{szegedy2013intriguing}
Christian Szegedy, Wojciech Zaremba, Ilya Sutskever, Joan Bruna, Dumitru Erhan,
  Ian Goodfellow, and Rob Fergus.
\newblock Intriguing properties of neural networks.
\newblock \emph{arXiv preprint arXiv:1312.6199}, 2013.

\bibitem[Tan and Le(2019)]{tan2019efficientnet}
Mingxing Tan and Quoc~V Le.
\newblock Efficientnet: Rethinking model scaling for convolutional neural
  networks.
\newblock \emph{ICML}, 2019.

\bibitem[VainF(2020)]{VainF_2019}
VainF.
\newblock {DeepLabv3Plus-Pytorch}, 2020.
\newblock URL \url{https://github.com/VainF/DeepLabV3Plus-Pytorch}.

\bibitem[Wang et~al.(2019)Wang, Chen, Xu, Liu, Loy, and Lin]{wang2019carafe}
Jiaqi Wang, Kai Chen, Rui Xu, Ziwei Liu, Chen~Change Loy, and Dahua Lin.
\newblock Carafe: Content-aware reassembly of features.
\newblock In \emph{ICCV}, 2019.

\bibitem[Wu and He(2018)]{wu-eccv2018}
Yuxin Wu and Kaiming He.
\newblock Group normalization.
\newblock In \emph{ECCV}, 2018.

\bibitem[Xie et~al.(2019)Xie, Wu, Maaten, Yuille, and He]{xie2019feature}
Cihang Xie, Yuxin Wu, Laurens van~der Maaten, Alan~L Yuille, and Kaiming He.
\newblock Feature denoising for improving adversarial robustness.
\newblock In \emph{CVPR}, 2019.

\bibitem[Ye et~al.(2019)Ye, Zhang, Yuen, and Chang]{cvpr19uel}
Mang Ye, Xu~Zhang, Pong~C Yuen, and Shih-Fu Chang.
\newblock Unsupervised embedding learning via invariant and spreading instance
  feature.
\newblock In \emph{CVPR}, 2019.

\bibitem[Zhang(2020)]{zhang2019making}
Richard Zhang.
\newblock Making convolutional networks shift-invariant again.
\newblock In \emph{ICML}, 2020.

\end{thebibliography}
\end{document}